\def\year{2022}\relax
\documentclass[letterpaper]{article} 
\usepackage{aaai22}  
\usepackage{times}  
\usepackage{helvet}  
\usepackage{courier}  
\usepackage[hyphens]{url}  
\usepackage{graphicx} 
\usepackage{natbib}  
\usepackage{caption} 
\DeclareCaptionStyle{ruled}{labelfont=normalfont,labelsep=colon,strut=off} 
\usepackage[utf8]{inputenc} 
\usepackage{booktabs}       
\usepackage{amsfonts}       
\usepackage{amsmath}
\usepackage{nicefrac}       
\usepackage{microtype}      
\usepackage{xcolor}         
\usepackage{xspace}
\usepackage{algpseudocode}
\usepackage{bbold}
\usepackage{bold-extra}
\newcommand{\BERTL}{BERT$_{\text Large}$\xspace}
\newcommand{\RoBERTaL}{RoBERTa$_{\text Large}$\xspace}
\newcommand{\CLINC}{\textsc{clinc}\xspace}
\newcommand{\BANKING}{\textsc{banking}\xspace}
\newcommand{\HWU}{\textsc{hwu}\xspace}

\newcommand{\systemname}{\textit{CAT}\xspace}

\title{Improved Text Classification via Contrastive Adversarial Training}

\author{%
    Lin Pan$^\dagger$, Chung-Wei Hang$^\dagger$, Avirup Sil$^\mathsection$, Saloni Potdar$^\dagger$\\
}
\affiliations{
    IBM Watson$^\dagger$\\
    IBM Research AI$^\mathsection$\\
    {\tt \{panl, hangc, avi, potdars\}@us.ibm.com}
}

\begin{document}

\maketitle

\begin{abstract}
  We propose a simple and general method to regularize the fine-tuning of Transformer-based encoders for text classification tasks. Specifically, during fine-tuning we generate adversarial examples by perturbing the word embedding matrix of the model and perform contrastive learning on clean and adversarial examples in order to teach the model to learn noise-invariant representations. By training on both clean and adversarial examples along with the additional contrastive objective, we observe consistent improvement over standard fine-tuning on clean examples. On several GLUE benchmark tasks, our fine-tuned \BERTL model outperforms \BERTL baseline by $1.7\%$ on average, and our fine-tuned \RoBERTaL improves over \RoBERTaL baseline by $1.3\%$. We additionally validate our method in different domains using three intent classification datasets, where our fine-tuned \RoBERTaL outperforms \RoBERTaL baseline by $1$--$2\%$ on average. For the challenging low-resource scenario, we train our system using half of the training data (per intent) in each of the three intent classification datasets, and achieve similar performance compared to the baseline trained with full training data.  
\end{abstract}

\section{Introduction}
Adversarial training (AT) introduced in \citet{Goodfellow+ICLR-15} provides an effective means of regularization and improving model robustness against adversarial examples \cite{szegedy+2014} for computer vision (CV) tasks such as image classification. In AT of this form, a small, gradient-based perturbation is added to the original example, and a model is trained on both clean and perturbed examples. Due to the discrete nature of textual data, this method is not directly applicable to NLP tasks. \citet{Miyato+ICLR-17} extend this method to NLP and propose to apply perturbation to the word embeddings of an LSTM-based model \cite{Hochreiter-Schmidhuber-LSTM-97} on text classification tasks. Since the word embeddings after perturbation do not map to new words in the vocabulary, the method is proposed exclusively as a means of regularization. 

In this work, we present \systemname, \underline{c}ontrastive \underline{a}dversarial \underline{t}raining for text classification. We build upon~\citet{Miyato+ICLR-17} to regularize the fine-tuning of Transformer-based \cite{Vaswani+17} encoders on text classification tasks. Instead of applying perturbation to word embeddings, we make a small change and apply it to the word embedding matrix of Transformer encoders and observe slightly better result in our experiments. Additionally, we encourage the model to learn noise-invariant representations by introducing a contrastive objective \cite{Oord+19} that pushes clean examples and their corresponding perturbed examples close to each other in the representation space, while pushing apart examples not from the same pair. We evaluate our method on a range of natural language understanding tasks including the standard GLUE \cite{Wang+GLUE-ICLR-19} benchmark as well as three intent classification tasks for dialog systems. 

On GLUE tasks, we compare our fine-tuning method against strong baselines of fine-tuning \BERTL \cite{Devlin+18} and \RoBERTaL \cite{Liu+19} on clean examples with the cross-entropy loss. Our method outperforms \BERTL by $1.7\%$ on average and \RoBERTaL by $1.3\%$. On intent classification tasks, our fine-tuned \RoBERTaL outperforms \RoBERTaL baseline by $1\%$ on the full test sets and $2\%$ on the difficult test sets. We further perform sample efficiency tests, where we use only half of the training data (per intent) and achieve near identical accuracy compared to the baseline trained using full training data.

\section{Related Work}

\subsection{Adversarial Training}
Adversarial Training (AT) has been explored in many supervised classification tasks that include object detection \cite{chen2018robust, song2018physical, xie2017adversarial}
, object segmentation \cite{arnab2018robustness, xie2017adversarial} and image classification \cite{Goodfellow+ICLR-15, papernot2016limitations, su2019one}. 

AT can be defined as the process in which a system is trained to defend against malicious ``attacks'' and increase network robustness,
by training the system with adversarial examples and optionally with clean examples. Typically these attacks are produced by perturbing the input (clean) examples, which makes the system predict the wrong class label \cite{chakraborty2018adversarial,
yuan2019adversarial}. In this paper, we limit our discussion to white-box adversarial attacks, i.e., assuming access to model architecture and parameters.

\citet{Miyato+ICLR-17} extend the \emph{Fast Gradient Sign Method} (FGSM) proposed in \citet{Goodfellow+ICLR-15} to NLP tasks by perturbing word embeddings, and applies the method to both supervised and semi-supervised settings with Virtual Adversarial Training (VAT) \cite{miyato2016distributional} for the latter. \cite{Wu+EMNLP-17} applies AT to relation extraction. Recent works \cite{kitada2020attention, Kitada-Iyatomi+21, Zhu+FreeLB-ICLR-20} propose to apply perturbations to the attention mechanism in Transformer-based encoders. Compared to single-step FGSM, \citet{Madry+ICLR-18} demonstrate the superior effectiveness of the multi-step approach to generate perturbed examples with projected gradient descent, which comes at a greater computational cost due to the inner loop that iteratively calculates the perturbations. \citet{shafahi2019adversarial} propose ``free'' adversarial training. In the inner loop where perturbations are calculated, gradients with respect to model parameters are also calculated and updated. The number of training epochs are also reduced to achieve comparable complexity with natural training. \citet{Zhu+FreeLB-ICLR-20} adopt the ``free'' AT algorithm and further add gradient accumulation to achieve a larger effective batch. Similar to \citet{Miyato+ICLR-17}, perturbations are applied to word embeddings of LSTM and BERT-based models. In our work, we use the simpler one-step FGSM to generate perturbed examples and perform contrastive learning with clean examples. 

\subsection{Contrastive Learning}
Recent advances in self-supervised contrastive learning, such as MoCo \cite{He+20} and SimCLR \cite{Chen+2020} have bridged the gap in performance between self-supervised learning and fully-supervised methods on the ImageNet \cite{Deng+09} dataset. Several works have successfully applied this representation learning paradigm to various NLP tasks. 

A key component in contrastive learning is how to create positive pairs. \cite{fang2020cert} uses back-translation to generate another view of the original English data. \citet{Wu+CLEAR-20} apply word and span deletion, reordering, and substitution. \citet{meng2021cocolm} use sequence cropping and masked sequence from an auxiliary Transformer. \citet{Giorgi+DeCLUTR-ACL-21} use nearby text spans in a document as positive pairs. \citet{Gunel+ICLR-21} treat training examples of the same class as positive pairs and performs supervised contrastive learning \cite{Khosla-NIPS-20}. \citet{Gao+SimCSE-21} use different dropout masks on the same batch data to generate positive pairs. As a supervised alternative, they leverage NLI datasets~\cite{bowman-etal-2015-large, Williams+18} and treat premises and their corresponding hypotheses as positive pairs and contradictions as hard negatives. 

In our work, we treat an original example, and its adversarial example as a positive pair, and a contrastive loss is used as an additional regularizer during fine-tuning. For multilingual NLP, \citet{Chi+20,pan-2021-multilingual,Wei+20} leverage parallel data and perform contrastive learning on parallel sentences for cross-lingual representation learning. 

\subsection{Improving BERT for Text Classification}
As a general method to improve Transformer-based model performance on downstream tasks, \citet{Sun+2020} and \citet{Gururangan+ACL-20} propose further language model pretraining in the target domain before the final fine-tuning. \citet{Du+20} propose to use self-training as another way to leverage unlabeled data, where a teacher model is first trained on labeled data, and is then used to label large amount of in-domain unlabeled data for the student model to learn from. 

Recent developments in language model pretraining have also advanced state-of-the-art results on a wide range of NLP tasks. ELECTRA \citet{Clark+ELECTRA-ICLR-20} use a generator to generate noisy text via the mask language modeling objective and a discriminator is used to classify each input token as \textit{original} or \textit{replaced}. The model shows strong performance on downstream tasks by fine-tuning the discriminator. DeBERTa \cite{He+DeBERTa-ICLR-21} proposes a disentangled attention mechanism and a new form of VAT, where perturbations are applied to the normalized word embeddings.

\begin{figure}[t!]
\centering
\includegraphics[width=.49\textwidth]{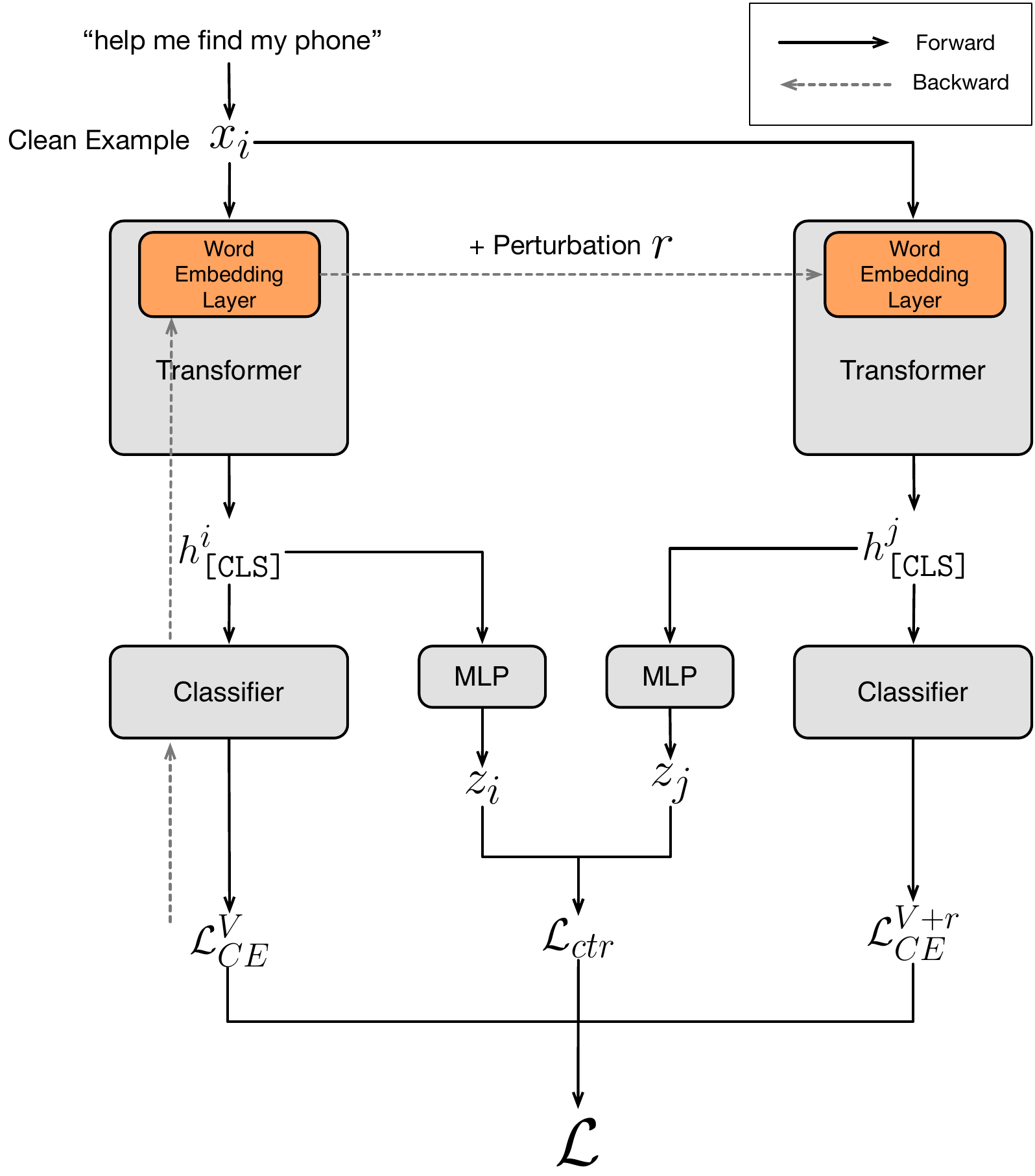}
\caption{Model architecture for our proposed method to fine-tune Transformer-based encoders on text classification tasks. We use the Fast Gradient Sign Method to generate adversarial examples by perturbing the word embedding matrix $V$ of the encoder. We then train on both clean and perturbed examples with the cross-entropy loss. Additionally, we introduce a third, contrastive loss that brings the representations of clean examples and their corresponding perturbed examples close to each other in order for the model to learn noise-invariant representations.}
\label{fig:method}
\end{figure}

\section{Method}
In this section we first briefly describe the standard fine-tuning procedure for Transformer-based encoders on text classification tasks. We then introduce our method of generating adversarial examples and propose our method \systemname{} that uses these examples to perform contrastive learning with clean examples. Figure~\ref{fig:method} shows our overall model architecture.

\subsection{Preliminaries}
Our learning setup is based on a standard multi-class classification problem with input training examples $\{x_i,y_i\}_{i=1,\ldots,N}$. We assume access to a Transformer-based pre-trained language model (PLM), such as BERT, and RoBERTa. 
Given a token sequence $x_i = [CLS, t_1, t_2,\ldots,t_T,SEP]$ \footnote{In the case of sequence pairs, another [SEP] token is added in between the sequences.}, the PLM outputs a sequence of contextualized token representations $H^L = [h_{[CLS]}^L,h_1^L,h_2^L,\ldots,h_T^L,h_{[SEP]}^L]$. 
\begin{eqnarray*}
\lefteqn{h_{[CLS]}^L, h_1^L,\ldots,h_T^L,h_{[SEP]}^L = } \\
& & PLM([CLS],t_1,\ldots,t_T,[SEP]),
\end{eqnarray*}
where $L$ denotes the number of model layers \footnote{We drop the layer superscript from here on for notation convenience.}.

The standard practice for fine-tuning these large PLMs is to add a softmax classifier on top of the model's sentence-level representations, such as the final hidden state $h_{[CLS]}$ of the $[CLS]$ token in BERT:

\vspace{1mm}
\begin{equation}
p(y_c | h_{[CLS]}) = \mathit{softmax}(W h_{[CLS]}) \;\;\; c \in C \label{eq1},
\end{equation}
\vspace{1mm}

where $W\in \mathbb{R}^{d_C\times d_h}$, and C denotes the number of classes. A model is trained by minimizing the cross entropy loss:

\vspace{1mm}
\begin{equation}
\mathcal{L}_{CE} = -\frac{1}{N} \sum_{i=1}^N \sum_{c=1}^C y_{i,c} \log(p(y_{i,c} | h_{[CLS]}^i)) \label{eq2},
\end{equation}
\vspace{1mm}

where $N$ is batch size.

\subsection{Adversarial Examples}
Adversarial examples are imperceptibly perturbed input to a model that causes misclassification. \citet{Goodfellow+ICLR-15} propose the \emph{Fast Gradient Sign Method} (FGSM) to generate such examples, and training on both clean and adversarial examples as an efficient way to improve model robustness against adversaries. 

Formally, given a loss function $\mathcal{L}(f_{\theta}(x_i+r), y_i)$, where $f_{\theta}$ is a neural network parameterized by $\theta$, and $x_i$ the input example, we maximize the loss function subject to the max norm constraint on the perturbation $r$:

\vspace{1mm}
\begin{equation}
\max_{r}\mathcal{L}(f_{\theta}(x_i+r), y_i), s.t. \left\Vert r \right\Vert_{\infty} < \epsilon \label{eq3},
\;\;\;\text{where $\epsilon > 0$} 
\end{equation}
\vspace{1mm}

Using first-order approximation, the loss function is approximately equivalent to the following:
\vspace{1mm}
\begin{equation}
\mathcal{L}(f_{\theta}(x_i+r), y_i) \approx \mathcal{L}(f_{\theta}(x_i), y_i) + \nabla_{x_i}\mathcal{L}(f_{\theta}(x_i), y_i)^{T}r \label{eq4}
\end{equation}
\vspace{1mm}

Solving for (\ref{eq3}) and (\ref{eq4}) yields perturbation in the following form:
\vspace{1mm}
\begin{equation}
r = -\epsilon \: \mathit{sign}(\nabla_{x_i}\mathcal{L}(f_{\theta}(x_i), y_i)) \label{eq5}
\end{equation}
\vspace{1mm}

Alternatively, using $l_2$-norm constraint on the perturbation $r$ in (\ref{eq3}) yields:
\vspace{1mm}
\begin{equation}
r = -\epsilon \frac{\nabla_{x_i}\mathcal{L}(f_{\theta}(x_i), y_i)} {\left\Vert \nabla_{x_i}\mathcal{L}(f_{\theta}(x_i), y_i)\right\Vert_{2}} \label{eq6}
\end{equation}
\vspace{1mm}

AT in \cite{Goodfellow+ICLR-15} uses both the clean example $x_i$ and perturbed example $x_i+r$ to train a model. For NLP problems where input is usually discrete, FGSM is not directly applicable. \citet{Miyato+ICLR-17} propose to apply perturbation to the word embedding $v_i$ from the corresponding row in embedding matrix $V \in \mathbb{R}^{d_v\times d_h}$, where $d_v$ is vocabulary size and $d_h$ hidden size. 

We follow this approach, but instead of perturbing the word embeddings, we directly perturb the \textit{word embedding matrix} of Transformer-based encoders to generate our adversarial examples. Specifically, after each forward pass with clean examples, we calculate the gradient of the loss function in (\ref{eq2}) with respect to the word embedding matrix $V$ instead of word embedding in (\ref{eq5}) to calculate the perturbation. Empirically, we find that perturbing the word embedding matrix performs better than word embeddings (see a comparison on GLUE in Table \ref{tab:exp-glue-embed-vs-V}). For text classification tasks, we train on clean and adversarial examples with the cross entropy loss in (\ref{eq2}). Additionally, we experiment with different forms of perturbation in (\ref{eq5}) and (\ref{eq6}), as well as randomly sampling between the two for each batch of data (\ref{tab:exp-glue-positive-pair-selection}). We observe that using $r$ with the max norm constraint consistently leads to the best result. In the Experiment section, we report results from using this form of perturbation.

\begin{table*}[ht!]
    \centering
    \scalebox{0.9}{%
    \begin{tabular}{l|lcrlrr}
    \toprule
    {\bf Dataset} & \multicolumn{1}{c}{\bf Task} & {\bf Labels} & {\bf Train} & {\bf Metric} & {\bf Train avg length} & {\bf Dev avg length} \\
    \midrule
    MNLI & Textual entailment & 3 & 393k & Accuracy & 29 & 28 \\
    QQP & Question paraphrase & 2 & 364k & Accuracy & 21 & 21 \\
    QNLI & Question answering/Textual entailment & 2 & 105k & Accuracy & 35 & 37 \\
    MRPC & Paraphrase & 2 & 3.k & F1 & 38 & 39 \\
    RTE & Textual entailment & 2 & 2.5k & Accuracy & 51 & 50 \\
    CoLA & Grammatical correctness & 2 & 8.5k & MCC & 8 & 8 \\
    SST-2 & Sentiment analysis & 2 & 67k & Accuracy & 9 & 17 \\
    \bottomrule
    \end{tabular}%
    }
    \vspace{2mm}
    \caption{GLUE sequence classification datasets statistics.}
    \label{tab:glue-datasets}
\end{table*}

\subsection{Contrastive Learning}
Intuitively, given a pair of clean and adversarial examples, we want their encoded sentence-level representation to be as similar to each other as possible so that our trained model will be more noise-invariant. At the same time, examples not from the same pair should be farther away in the representation space. 

To model this relationship, we leverage contrastive learning as an additional regularizer during the fine-tuning process. Recent works on contrastive learning, such as MoCo \cite{He+20}, and SimCLR \cite{Chen+2020} use various forms of data augmentation, e.g., random cropping, and random color distortion, as the first step to create positive pairs. MoCo uses a queue structure to store negative examples, while SimCLR performs in-batch negative example sampling. A model is then trained by minimizing the InfoNCE loss. In our work, we employ the SimCLR formulation of positive and negative pairs, and its loss function to implement our contrastive objective. 

Concretely, given the final hidden state $h_{[CLS]}^i$ of the $[CLS]$ token for a clean example, and $h_{[CLS]}^j$ for its corresponding adversarial example, we treat ($h_{[CLS]}^i$, $h_{[CLS]}^j$) as a pair of positive examples. Following \citet{Chen+2020}, we add a non-linear projection layer on top of them:

\vspace{1mm}
\begin{align}
z_i &= W_2 \mathit{ReLU} (W_1 h_{[CLS]}^i), \label{eq7} \\
z_j &= W_2 \mathit{ReLU} (W_1 h_{[CLS]}^j) \label{eq8}
\end{align}
\vspace{1mm}

where $W_1\in \mathbb{R}^{d_h\times d_h}$, $W_2\in \mathbb{R}^{d_k\times d_h}$, and $d_k$ is set to $300$. With a batch of $N$ clean examples and their corresponding adversarial examples, for each positive pair, there are $2(N - 1)$ negative pairs, i.e., all the rest of the examples in the batch are negative examples. The contrastive objective is to identify the positive pair:
\vspace{1mm}
\begin{equation}
\mathcal{L_{\text{ctr}}} = -\log \frac{\exp(sim(z_i, z_j /\tau))}{\sum_{k=1}^{2N}\mathbb{1}_{[k\neq i]} \exp(sim(z_i, z_k /\tau))}, \label{eq9}
\end{equation}
\vspace{1mm}

where $\mathit{sim(u, v)} = u^Tv/ \left\Vert u \right\Vert_{2} \left\Vert v \right\Vert_{2}$ denotes cosine similarity between two vectors, and $\tau$ a temperature hyperparameter.

Finally, we perform fine-tuning in a multi-task manner and take a weighted average of the two classification losses and the contrastive loss \footnote{In our experiments, we always assign equal weights to the two classification losses. It is possible that a different weight distribution yields better results.}:
\vspace{1mm}
\begin{equation}
\mathcal{L} = \frac{(1 - \lambda)}{2} (\mathcal{L}_{CE}^V + \mathcal{L}_{CE}^{V+r}) + \lambda \mathcal{L}_{ctr} \label{eq10}
\end{equation}
\vspace{1mm}

            

\section{Experiment} \label{Exp}
\subsection{Datasets}
We conduct experiments on seven tasks of the GLUE benchmark, including textual entailment (MNLI, RTE), question answering/entailment (QNLI), question paraphrase (QQP), paraphrase (MRPC), grammatical correctness (CoLA), and sentiment analysis (SST-2). Table~\ref{tab:glue-datasets} summarizes the statistics of the GLUE tasks.

\begin{table}[t!]
    \centering
    \scalebox{0.8}{%
    \begin{tabular}{l|rr|rrr|r}
    \toprule
    & & & \multicolumn{3}{c|}{\bf Examples} \\
    {\bf Dataset} & {\bf Intents} & {\bf Domains} & {\bf Train} & {\bf Test} & {\bf Test} & {\bf Avg len}\\
    & & & & & (difficult) \\
    \midrule
    \CLINC & 150 & 10 & 17,999 & 4,500 & 750 & 8 \\
    \BANKING & 77 & 1 & 10,003 & 3,080 & 770 & 12 \\
    \HWU & 64 & 21 & 9,957 & 1,076 & 620 & 7\\
    \bottomrule
    \end{tabular}%
    }
    \vspace{2mm}
    \caption{Intent classification dataset statistics.}
    \label{tab:intent-datasets}
\end{table}

We additionally experiment on three commonly used intent classification datasets---\CLINC \cite{Larson+CLINC150-EMNLP-19}, \BANKING \cite{Casanueva+Banking77-20} and \HWU~\cite{Liu+HWU64-IWSDS-19}. Intent classification is the process of identifying the class (intent) of any utterance in a task-oriented dialog system. These three datasets largely represent a short-text classification task in real-world settings. Table \ref{tab:intent-datasets} summarizes the statistics of the three datasets. 

\begin{itemize}
\item {\textbf{\CLINC}} covers $150$ intents in $10$ domains (e.g., banking, work, auto, travel). The dataset is designed to capture the breadth of topics that a production task-oriented chatbot handles. The dataset also comes with $1,200$ out-of-scope examples. In this work, we focus on in-scope examples only.
\item {\textbf{\BANKING}} is a single domain dataset created for fine-grained intent classification. The dataset consists of customer service queries in the banking domain, covering $77$ intents across $10,003$ training examples and $3080$ test examples.
\item {\textbf{\HWU}} covers $64$ intents in $21$ domains (e.g., alarm, email, game, news). The dataset is created in the context of a real-world use case of a home assistant bot. We use the one fold train-test split with $9,957$ training examples and $1,076$ test examples for our experiments.
\end{itemize}

\begin{table*}[ht!]
\centering
\scalebox{.999}{%
\begin{tabular}{l|lllllll|c}
\toprule
{\bf Model} & \multicolumn{1}{c}{\bf MNLI} & \multicolumn{1}{c}{\bf QQP} &  \multicolumn{1}{c}{\bf QNLI} & \multicolumn{1}{c}{\bf MRPC} & \multicolumn{1}{c}{\bf RTE} & \multicolumn{1}{c}{\bf CoLA} & \multicolumn{1}{c|}{\bf SST-2} & \multicolumn{1}{c}{\bf Avg} \\
\midrule
\BERTL (our impl)                 & 86.6 & 91.4 & 92.0 & 90.0 & 69.3 & 61.8 & 93.6 & 83.5 \\
\BERTL + AT + CTR (ours)                    & \underline{87.4}$\dagger$ & \underline{92.2}$\dagger$ & \underline{93.0}$\dagger$ & \underline{91.6} & \underline{71.5} & \underline{65.8}$\dagger$ & \underline{95.2}$\dagger$ & \underline{85.2} \\
\midrule
\RoBERTaL \cite{Liu+19} & 90.2 & 92.2 & 94.7 & 90.9 & 86.6 & 68.0 & 96.4 & 88.4 \\
\RoBERTaL (our impl)              & 90.5 & 91.8 & 94.5 & 90.6 & 85.9 & 67.0 & 96.1 & 88.1 \\
\RoBERTaL + AT + CTR (ours)                 & \bf{91.1}$\dagger$ & 92.5$\dagger$ & \bf{95.1}$\dagger$ & \bf{93.0}$\dagger$ & 87.4 & 69.4 & \bf{97.0} & \bf{89.4} \\
\midrule
SCL \cite{Gunel+ICLR-21} & 88.6 & \multicolumn{1}{c}{-} & 93.9 & 89.5 & 85.7 & 86.1* & 96.3 & - \\
PGD from \cite{Zhu+FreeLB-ICLR-20} & 90.5 & 92.5 & 94.9 & 90.9 & 87.4 & 69.7 & 96.4 & 88.9 \\
FreeAT from \cite{Zhu+FreeLB-ICLR-20} & 90.0 & 92.5 & 94.7 & 90.7 & 86.7 & 68.8 & 96.1 & 88.5 \\
FreeLB \cite{Zhu+FreeLB-ICLR-20} & 90.6 & \bf{92.6} & 95.0 & 91.4 & \bf{88.1} & \bf{71.1} & 96.8 & 89.4 \\ 
\bottomrule
\end{tabular}%
}
\caption{Results on the dev sets of GLUE benchmark. \textit{AT} refers to the adversarial training component of our system and \textit{CTR} the contrastive learning component. On average, our fine-tuned \BERTL model outperforms \BERTL baseline by $1.7\%$, and our fine-tuned \RoBERTaL improves over \RoBERTaL baseline by $1.3\%$. We fine-tune \BERTL and \RoBERTaL from their original checkpoints with no task-wise transfer learning involved. $\dagger$ indicates statistically significant improvement over the baseline. For CoLA and MRPC, we use Fisher Randomization test, and McNemar's test for all the other tasks. *It is unclear to us how the result for CoLA was derived in \cite{Gunel+ICLR-21} since their baseline \RoBERTaL is significantly higher than the ones reported in \cite{Liu+19} and other related works.} 
\label{tab:exp-glue}
\end{table*}

On average, the sentence length of intent classification datasets is shorter than that of the GLUE tasks \footnote{Sentence length is measured by the number of words instead of tokens since different BERT-like models use different tokenizers.}, since most of the GLUE tasks consist of two sentences. However, the number of classes is much greater (Table~\ref{tab:glue-datasets} and \ref{tab:intent-datasets}).

For the three intent classification datasets, in addition to the original evaluation data, we also evaluate on a \textit{difficult} subset of each test set described in \cite{qi+2021}. The difficult subsets are constructed by comparing the TF-IDF vector of each test example to that of the training examples for a given intent. The test examples that are most dissimilar to the corresponding training examples are selected for inclusion to the difficult subset. We also experimented with generating our own difficult subsets in a similar manner using BERT-based sentence encoders \footnote{The specific model used is ``paraphrase-mpnet-base-v2'' \cite{song2020mpnet}, available from \url{https://www.sbert.net/docs/pretrained_models.html.}}, and compare each test example with the mean-pooling of the training examples for that intent. Result shows that the TF-IDF method yields a more challenging subset, thus we report results on the original subsets from \citet{qi+2021}. The evaluation metric for all intent classification datasets is accuracy.

\subsection{Training Details}
We apply \systemname\ to the fine-tuning of two backbone PLMs, \BERTL and \RoBERTaL.
For all experiments, we use AdamW optimizer with $0.01$ weight decay and a linear learning rate scheduler. We set max sequence length to $128$ and learning rate warmup for the first $10\%$ of the total iterations. 

On GLUE tasks, we largely follow the hyperparameter settings reported in \citet{Devlin+18} and \citet{Liu+19} \footnote{For \RoBERTaL baseline, we follow the hyperparameter settings specified at \url{https://github.com/pytorch/fairseq/blob/master/examples/roberta/README.glue.md}} to generate our \BERTL and \RoBERTaL baselines. For \BERTL, we set batch size to $32$ and fine-tune for $3$ epochs. Grid search is performed over $lr \in \{0.00001, 0.00002, 0.00003\}$. For \RoBERTaL, we sweep over the same learning rates as \BERTL and batch size $\in \{16, 32\}$. 

On the three intent classification datasets, we follow the hyperparameter settings in \citet{qi+2021}. We use a batch size of $32$ and fine-tune for $5$ epochs, and search over $lr \in \{0.00003, 0.00004, 0.00005\}$. 

For fine-tuning with \systemname, we use the exact same hyperparameter settings as the baseline, and further perform grid search over $\epsilon \in \{0.0001, 0.001, 0.005, 0.02\}$, $\tau \in \{0.05, 0.06, 0.07, 0.08, 0.09, 0.1\}$, and $\lambda \in \{0.1, 0.2, 0.3, 0.4, 0.5\}$. All our experiments were run on a single 32 GB V100 GPU.

\begin{table}[t!]
\centering
\begin{tabular}{l|lll|c}
\toprule
{\bf Model} & \multicolumn{1}{c}{\bf\small \CLINC} & \multicolumn{1}{c}{\bf\small \BANKING} & \multicolumn{1}{c}{\bf\small \HWU} & \multicolumn{1}{|c}{\bf\small Avg} \\
\midrule
RoBERTa$_L$         & 97.4 & 93.9 & 92.4 & 94.6 \\
RoBERTa$_L$+AT+CTR & \bf{98.0}$\dagger$ & \bf{95.0}$\dagger$ & \bf{93.8}$\dagger$ & \bf{95.6} \\
\bottomrule
\end{tabular}
\vspace{2mm} 
\caption{Results on full test sets of intent classification datasets. \textit{AT} refers to the adversarial training component of our system and \textit{CTR} the contrastive learning component. On average, our fine-tuned \RoBERTaL improves over \RoBERTaL baseline by $1\%$. $\dagger$ indicates statistically significant improvement over the baseline using McNemar's test.}
\label{tab:exp-intent}
\end{table}

\begin{table}[t!]
\centering
\begin{tabular}{l|lll|c}
\toprule
{\bf Model} & \multicolumn{1}{c}{\bf\small \CLINC} & \multicolumn{1}{c}{\bf\small \BANKING} & \multicolumn{1}{c}{\bf\small \HWU} & \multicolumn{1}{|c}{\bf\small Avg} \\
\midrule
RoBERTa$_L$         & 91.1 & 83.8 & 89.5 & 88.1 \\
RoBERTa$_L$+AT+CTR & \bf{92.1}$\dagger$ & \bf{87.3}$\dagger$ & \bf{90.8}$\dagger$ & \bf{90.1} \\
\bottomrule
\end{tabular}
\vspace{2mm}
\caption{Results on \textit{difficult} test sets of intent classification datasets. On average, our fine-tuned \RoBERTaL improves over \RoBERTaL baseline by $2\%$. The largest improvement is made on the \BANKING dataset with $3.5\%$. $\dagger$ indicates statistically significant improvement over the baseline using McNemar's test.}
\label{tab:exp-intent-tfidf}
\end{table}

\begin{table*}[!t]
\centering
\begin{tabular}{l|c|ccc|c}
\toprule
{\bf Model} & \bf Training data & \bf \CLINC & \bf \BANKING & \bf \HWU & \bf Avg \\
\midrule
\RoBERTaL         & \bf full & \bf{97.4} & 93.9 & \bf{92.4} & \bf{94.6} \\
\RoBERTaL + AT + CTR & \bf half & 97.1 & \bf{94.2} & 92.3 & 94.5 \\
\bottomrule
\end{tabular}
\vspace{2mm}
\caption{Sample efficiency test results on the full test sets of intent classification datasets. With our proposed fine-tuning method, we can use about half of the training data (per intent) and achieve near identical accuracy compared to baseline trained with full training data.}
\label{tab:exp-intent-efficiency}
\end{table*}

\begin{table*}[!ht]
\centering
\scalebox{1.}{%
\begin{tabular}{l|ccccccc|c}
\toprule
{\bf Model} & \bf MNLI & \bf QQP & \bf QNLI & \bf MRPC & \bf RTE & \bf CoLA & \bf SST-2 & \bf Avg \\
\midrule
\RoBERTaL                             & 90.5 & 91.8 & 94.5 & 90.6 & 85.9 & 67.0 & 96.1 & 88.1 \\
\RoBERTaL + AT                       & 90.7 & 92.1 & 94.9 & 92.4 & 85.6 & \bf{69.9} & 96.6 & 88.9 \\
\RoBERTaL + AT + CTR                 & \bf{91.1} & \bf{92.5} & \bf{95.1} & \bf{93.0} & \bf{87.4} & 69.4 & \bf{97.0} & \bf{89.4} \\
\bottomrule
\end{tabular}%
}
\vspace{2mm}
\caption{Ablation results on the dev sets of GLUE benchmark. Adding our proposed AT that applies perturbation to the word embedding matrix leads to an improvement of $0.8\%$ over the baseline, achieving the same performance as PGD in Table~ \ref{tab:exp-glue}. Further adding the contrastive objective contributes to an additional $0.5\%$ improvement.}
\label{tab:exp-glue-abl}
\end{table*}

\subsection{GLUE Results}
On GLUE tasks, we fine-tune \BERTL and \RoBERTaL using our method with two classification losses (on clean examples, and adversarial examples, respectively) and the contrastive loss. We compare them with the \BERTL and \RoBERTaL baseline, which are conventionally fine-tuned with classification loss on clean examples. For all experiments, we fine-tune \BERTL and \RoBERTaL from their original checkpoints with no task-wise transfer learning involved. We accompany each set of experiments with statistical significance test. For tasks evaluated with accuracy, we use McNemar's test. For CoLA, which is evaluated with Matthews correlation coefficient (MCC), and MRPC with F1, we use Fisher Randomization test. 

Table~\ref{tab:exp-glue} shows the dev set results.
In summary, \systemname\ fine-tuning approach consistently outperforms the standard fine-tuning approach for both \BERTL and \RoBERTaL.
\systemname\ leads to a $1.7\%$ improvement on average over conventionally fine-tuned \BERTL.
The largest improvement is observed on the CoLA task (i.e., $4.0\%$).
For the stronger baseline of \RoBERTaL, we observe an improvement of $1.3\%$ on average with our method. On MRPC and CoLA, our result improves over \RoBERTaL baseline by $2.4\%$, showing the effectiveness our method on both single sequence, as well as sequence pair classification tasks.
For statistical significance, we note that our improved results over the baseline are not significant on some smaller datasets, e.g., improved \BERTL on MRPC and RTE, and \RoBERTaL on MRPC, RTE and CoLA.

We also list results for three other AT methods: PGD \cite{Madry+ICLR-18}, FreeAT \cite{shafahi2019adversarial} and FreeLB \cite{Zhu+FreeLB-ICLR-20}, as well as \citet{Gunel+ICLR-21} that use supervised contrastive learning for text classification. The three AT methods, which calculate perturbations iteratively, have been shown to produce stronger attacks compared to single-step methods \cite{Athalye+ICML-19}. Our single-step FGSM with perturbation applied to the word embedding matrix together with contrastive learning outperforms PGD and FreeAT while achieving the same overall performance as FreeLB. We note that baseline results from different papers are slightly different, which affects the results of proposed methods.  

\vspace{2mm}
\begin{table}[t!]
\centering
\begin{tabular}{l|ccc|c}
\toprule
{\bf Model} & \multicolumn{1}{c}{\bf\small \CLINC} & \multicolumn{1}{c}{\bf\small \BANKING} & \multicolumn{1}{c}{\bf\small \HWU} & \multicolumn{1}{|c}{\bf\small Avg} \\
\midrule
RoBERTa$_L$         & 91.1 & 83.8 & 89.5 & 88.1 \\
RoBERTa$_L$+AT   & 90.8 & 85.8 & 89.8 & 88.8 \\
RoBERTa$_L$+AT+CTR & \bf{92.1} & \bf{87.3} & \bf{90.8} & \bf{90.1} \\
\bottomrule
\end{tabular}
\vspace{2mm}
\caption{Ablation results on \textit{difficult} test sets of intent classification datasets. Adding AT improves over the baseline by $0.7\%$. Further adding the contrastive objective contributes to an additional $1.3\%$ improvement.}
\label{tab:exp-intent-tfidf-abl}
\end{table}

\begin{table*}[ht!]
\centering
\scalebox{1.}{%
\begin{tabular}{l|lllllll|c}
\toprule
{\bf Model} & \multicolumn{1}{c}{\bf MNLI} & \multicolumn{1}{c}{\bf QQP} &  \multicolumn{1}{c}{\bf QNLI} & \multicolumn{1}{c}{\bf MRPC} & \multicolumn{1}{c}{\bf RTE} & \multicolumn{1}{c}{\bf CoLA} & \multicolumn{1}{c|}{\bf SST-2} & \multicolumn{1}{c}{\bf Avg} \\
\midrule
\BERTL + $AT^x$ + CTR                 & 87.1 & 91.9 & 92.8 & 91.3 & \underline{71.8} & 63.3 & 94.2 & 84.6 \\
\BERTL + $AT^V$ + CTR                    & \underline{87.4} & \underline{92.2} & \underline{93.0} & \underline{91.6} & 71.5 & \underline{65.8} & \underline{95.2} & \underline{85.2} \\
\RoBERTaL $AT^x$ + CTR              & 90.9 & 92.4 & 94.9 & 92.6 & 86.3 & 69.4 & 96.6 & 89.0 \\
\RoBERTaL + $AT^V$ + CTR                 & \bf{91.1} & \bf{92.5} & \bf{95.1} & \bf{93.0} & \bf{87.4} & \bf{69.4} & \bf{97.0} & \bf{89.4} \\
\bottomrule
\end{tabular}%
}
\vspace{2mm}
\caption{Comparison between perturbation applied to word embeddings and word embedding matrix of \BERTL and \RoBERTaL during \systemname fine-tuning. Results are on the dev sets of the GLUE benchmark. We use $AT^x$ to denote adversarial training with perturbation applied to word embeddings and $AT^V$ to the word embedding matrix. \textit{CTR} refers to the contrastive learning component. On average, perturbing word embedding matrix outperforms perturbing word embeddings by $0.6\%$ for \BERTL and $0.4\%$ for \RoBERTaL.}
\label{tab:exp-glue-embed-vs-V}
\end{table*}
\vspace{2mm}

\vspace{3mm}
\begin{table*}[!t]
\centering
\scalebox{1.}{%
\begin{tabular}{l|cccccc|c}
\toprule
{\bf Model} & \bf MNLI & \bf QNLI & \bf MRPC & \bf RTE & \bf CoLA & \bf SST-2 & \bf Avg \\
\midrule
\RoBERTaL + AT + CTR (max norm/$l_2$-norm)                      & 90.6 & 94.9 & 92.7 & 86.3 & 68.8 & 96.8 & 88.4 \\
\RoBERTaL + AT + CTR (max norm)                 & \bf{91.1} & \bf{95.1} & \bf{93.0} & \bf{87.4} & \bf{69.4} & \bf{97.0} & \bf{88.8} \\
\bottomrule
\end{tabular}%
}
\vspace{2mm}
\caption{Comparison between consistently using the max norm constraint in generating adversarial examples, and randomly selecting between max norm and $l_2$-norm for each batch of data. Results are on the dev sets of GLUE benchmark. Consistently using the max norm constraint is $0.4\%$ better than randomly selecting between max norm and $l_2$-norm.}
\label{tab:exp-glue-positive-pair-selection}
\end{table*}

\subsection{Intent Classification Results}
On the three intent classification datasets, \CLINC, \BANKING, and \HWU, we experiment with the stronger \RoBERTaL baseline. 
Table~\ref{tab:exp-intent} summarizes the results.
On average, \systemname\ outperforms the fine-tuned \RoBERTaL baseline by $1\%$, when evaluated on the full test sets of the three datasets. The largest improvement is made on the \HWU dataset ($1.4\%$). 

Furthermore, we show that by training with adversarial examples and contrastive learning, \systemname\ makes \RoBERTaL work better on the difficult test subsets of the three intent classification tasks.
As shown in Table~\ref{tab:exp-intent-tfidf},
our approach improves over standard \RoBERTaL fine-tuning by $2\%$ on average. On \BANKING, our method results in a large improvement of $3.5\%$. For statistical significance test on intent classification datasets, we use McNemar's test and observe significant improvement over baseline results for all datasets and evaluation settings.

\subsection{Sample Efficiency}
Next, we demonstrate that \systemname\ has better sample efficiency compared to standard fine-tuning.
We design the experiment on the three intent classification datasets.
Specifically, we use about half of the training data (per intent) from each dataset to fine-tune \RoBERTaL with \systemname. The training examples are randomly sampled from the full training data.

As Table~\ref{tab:exp-intent-efficiency} shows, with the usage of only half of the training data, our method achieves nearly the same results, compared to standard fine-tuning using all the training data.
This result confirms the sample efficiency of our proposed \systemname, and also indicates the advantage of our approach in the challenging low-resource scenarios.

\subsection{Ablation}
Finally, we perform ablation experiments on both GLUE benchmark and the three intent classification datasets. Results are presented in Table \ref{tab:exp-glue-abl} and Table \ref{tab:exp-intent-tfidf-abl}. On GLUE, by removing the contrastive loss, i.e., fine-tuning with clean and adversarial examples, we observe an accuracy drop of $0.5\%$ on average. Compared to the baseline, this setup makes an average improvement of $0.8\%$. Our full system performs the best on all tasks except for CoLA, on which removing the contrastive loss yields the best performance. One interesting observation is that using our proposed perturbation to the word embedding matrix (instead of word embeddings), our \textit{\RoBERTaL+ AT} setting achieves the same average \mbox{accuracy} as PGD in Table~\ref{tab:exp-glue} while using the simpler single-step FGSM.

On intent classification datasets, we use the difficult test sets for evaluation. Here, we observe a much larger effect from the additional contrastive loss, improving over \textit{\RoBERTaL+ AT} by $1.3\%$ on average, while AT alone improves over the baseline by $0.7\%$. For all ablation experiments with the \textit{\RoBERTaL+ AT} setup, we perform a grid search over $\epsilon \in \{0.0001, 0.001, 0.005, 0.02\}$.

\section{Conclusion}
In this paper, we describe \systemname, a simple and effective method for regularizing the fine-tuning of Transformer-based encoders. By leveraging adversarial training and contrastive learning, our system consistently outperforms the standard fine-tuning method for text classification. We use strong baseline models and evaluate our method on a range of GLUE benchmark tasks and three intent classification datasets in different settings. Sample efficiency and ablation tests show the positive effects of combining our adversarial and contrastive objectives for improved text classification. In the future, we plan to study additional word-level objectives to complement the sentence-level contrastive learning objective, in order to extend our method to other NLP tasks.

\bibliography{chang}



\end{document}